\documentclass[10pt,twocolumn,letterpaper]{article}

\usepackage{cvpr}
\usepackage{graphicx}
\usepackage{amsmath}
\usepackage{amssymb}
\usepackage{booktabs}
\usepackage{hyperref}
\usepackage{cleveref}

\hypersetup{
    colorlinks=true,
    linkcolor=black,
    citecolor=black,
    urlcolor=black,
    linkbordercolor={1 1 1},
    citebordercolor={1 1 1},
    urlbordercolor={1 1 1},
}

\def\cvprPaperID{*****} 
\def\confName{CVPR}

\begin{document}

\title{Analysis of Premature Death Rates in Texas Counties:\\ The Impact of Air Quality, Socioeconomic Factors, and COPD Prevalence}

\author{Richard Rich\\
Texas A\&M University-San Antonio\\
{\tt\small rrich010@jaguar.tamu.edu}
\and
Ernesto Diaz\\
Texas A\&M University-San Antonio\\
{\tt\small ediaz029@jaguar.tamu.edu}
}

\maketitle

\begin{abstract}
Understanding factors contributing to premature mortality is critical for public health planning. This study examines the relationships between premature death rates and multiple risk factors across several Texas counties, utilizing EPA air quality data, Census information, and county health records from recent years. We analyze the impact of air quality (PM2.5 levels), socioeconomic factors (median household income), and health conditions (COPD prevalence) through statistical analysis and modeling techniques. Results reveal COPD prevalence as a strong predictor of premature death rates, with higher prevalence associated with a substantial increase in years of potential life lost. While socioeconomic factors show a significant negative correlation, air quality demonstrates more complex indirect relationships. These findings emphasize the need for integrated public health interventions that prioritize key health conditions while addressing underlying socioeconomic disparities.
\end{abstract}

\section{Introduction}
Premature death, measured by Years of Potential Life Lost (YPLL), is a crucial public health metric that quantifies the impact of early mortality on communities. Environmental factors such as air pollution, socioeconomic conditions including household income, and prevalent health conditions like Chronic Obstructive Pulmonary Disease (COPD) contribute to premature mortality rates. Understanding these complex relationships is essential for developing effective public health strategies. This study focuses on Texas counties, integrating environmental, health, and census datasets to identify and analyze key predictors of premature death rates. By examining these relationships, we aim to provide insights that can guide targeted public health interventions and resource allocation.

\section{Related Work}
Recent research has highlighted the complex relationships between environmental factors, socioeconomic conditions, and public health outcomes. A comprehensive study on the burden of cause-specific mortality by \cite{bowe2019burden} demonstrated that PM2.5 exposure is associated with increased mortality rates, with disproportionate effects on socioeconomically disadvantaged communities. Their findings revealed that 99\% of PM2.5-associated deaths occurred at levels below current EPA standards, emphasizing the need for continued investigation of air quality impacts even at lower concentration levels.

The methodological framework for analyzing such complex relationships has been extensively studied. \cite{chao2008quantifying} evaluated various methods for quantifying the relative importance of predictors in multiple regression analysis for public health studies, establishing best practices for assessing how different variables contribute to health outcomes. Building on these methodological insights, our study employs linear regression to examine the relative contributions of environmental and socioeconomic factors to premature death rates.

In examining broader analytical approaches to environmental health studies, \cite{agier2016systematic} conducted a systematic comparison of linear regression-based statistical methods to assess exposome-health associations. Their work highlighted the challenges of handling correlated environmental variables and the importance of selecting appropriate analytical methods. While they explored advanced statistical techniques, our study employs a focused linear regression approach to maintain interpretability while examining the specific context of Texas counties.

Previous spatial health analysis work by \cite{mollalo2020predicting} integrated GIS, spatial statistics, and machine learning algorithms to analyze respiratory infection mortality patterns nationwide. Their research predicting hotspots of age-adjusted mortality rates revealed significant relationships between socioeconomic factors and respiratory health outcomes. Our study builds upon this foundation by conducting a detailed county-level analysis in Texas, examining the specific interactions between PM2.5 levels, socioeconomic factors, and mortality rates.

This body of prior research establishes both the importance of investigating environmental and socioeconomic factors in public health outcomes and the methodological rigor required for such analysis. Our study contributes to this literature by providing a focused analysis of Texas counties, offering insights into the local manifestations of these broader national patterns.

\section{Methodology}
\subsection{Data Collection}
To investigate factors contributing to premature mortality in Texas, this study employed a rigorous data collection methodology, integrating information from multiple reliable sources. This sampling approach ensured a comprehensive and robust analysis while maintaining a high standard of data quality.

Air quality data, with a focus on PM2.5 concentrations, was acquired from the EPA's Air Quality System (AQS) database for the year 2022. This database provided standardized and reliable measurements of air quality, enabling consistent comparisons across the selected counties.

Socioeconomic data was sourced from the U.S. Census Bureau's American Community Survey 5-Year Estimates (2019). Median household income was selected as a key economic indicator, offering insights into the economic well-being of the studied communities.

To capture health outcomes, two primary data sources were utilized, both from the Centers for Disease Control and Prevention (CDC). Premature death rates, quantified as Years of Potential Life Lost (YPLL), were obtained from the County Health Rankings database. This measure provides a standardized assessment of premature mortality across counties. Additionally, COPD prevalence data was gathered, offering a direct measure of the burden of this respiratory condition in each county.

The collected data underwent several preprocessing steps to ensure analysis quality. County names were standardized across all datasets to enable accurate merging. This resulted in a final dataset of 29 counties with complete information across all variables. Numerical variables were scaled to appropriate units, with income measured in dollars, PM2.5 in µg/m³, COPD in percentage points, and YPLL in years of life lost.

The selection of features for our analysis was based on established public health literature and data availability. Median household income was chosen as a socioeconomic indicator due to its robust relationship with health outcomes. PM2.5 was selected as our air quality metric given its well-documented health impacts. COPD prevalence was included as a key health condition indicator due to its significant relationship with both air quality and mortality.

Multiple linear regression was chosen as our primary modeling approach after careful consideration of various statistical methods. While advanced techniques exist for handling multiple correlated exposures in environmental health studies \cite{agier2016systematic}, we opted for linear regression due to its interpretability and ability to quantify the relative importance of different predictors while controlling for other factors. This model type allows for direct interpretation of coefficients as the impact of each variable on premature death rates. The model's assumptions were verified through residual analysis and diagnostic plots, following established statistical practices in health sciences research.

Through the integration of diverse data sources, rigorous preprocessing, and strategic modeling approaches, this study established a comprehensive framework for analyzing the socioeconomic, environmental, and health factors influencing premature mortality across Texas counties.
\section{Data Sources and Preparation}

\subsection{Socioeconomic Data}
\textbf{Source:} U.S. Census Bureau ACS 5-Year Estimates.\\
\indent \textbf{Feature:} Median Household Income.\\
\indent \textbf{Processing:} Standardized county names and addressed missing values.

\subsection{Air Quality Data}
\textbf{Source:} EPA's Air Quality System (AQS).\\
\indent \textbf{Feature:} PM2.5 concentration (annual average).\\
\indent \textbf{Processing:} Filtered for Texas counties and aggregated data.

\subsection{Health Data}
\textbf{Source:} County Health Rankings.\\
\indent \textbf{Feature:} Premature Death Rate (YPLL).\\
\indent \textbf{Processing:} Standardized county names, converted rates to numeric, and removed missing values.

\subsection{COPD Prevalence Data}
\textbf{Source:} County-level COPD prevalence data.\\
\indent \textbf{Feature:} COPD prevalence (\%).\\
\indent \textbf{Processing:} Filtered for Texas counties, handled missing values, and standardized names.

\subsection{Data Merging}
Datasets were merged on county names to create a comprehensive dataset including:
\begin{itemize}
    \item Premature Death Rate (YPLL)
    \item Median Household Income
    \item PM2.5 Levels
    \item COPD Prevalence
\end{itemize}

\section{Experiments}

\subsection{Descriptive Statistics}

The descriptive statistics reveal considerable variation across Texas counties. Premature death rates range from 4,910 to 13,005 years of potential life lost, with a mean of 9,058 years. Median household income shows substantial economic disparity, ranging from \$38,758 to \$86,913, with an average of \$56,812. PM2.5 concentrations vary from 5.57 to 10.36 µg/m³, while COPD prevalence ranges from 4.7\% to 8.4\% across counties.

\begin{table}[ht]
\centering
\caption{Summary Statistics of Key Variables}
\label{table:descriptive_stats}
\begin{tabular}{lccc}
\toprule
\textbf{Variable} & \textbf{Mean} & \textbf{Min} & \textbf{Max} \\
\midrule
Premature Death Rate (YPLL) & 9,058.03 & 4,910 & 13,005 \\
Median Household Income (\$) & 56,812.83 & 38,758 & 86,913 \\
Average PM2.5 (\(\mu g/m^3\)) & 8.40 & 5.57 & 10.36 \\
COPD Prevalence (\%) & 6.39 & 4.7 & 8.4 \\
\bottomrule
\end{tabular}
\end{table}

\subsection{Correlation Analysis}

The correlation matrix reveals key relationships between variables. COPD prevalence shows the strongest correlation with premature death rates (0.76), indicating a robust positive relationship. Median household income demonstrates a moderate negative correlation (-0.51) with premature death rates, suggesting a protective effect. PM2.5 levels show unexpectedly weak correlations with both COPD prevalence (-0.017) and premature death rates (-0.17).

\begin{figure}[ht]
\centering
\includegraphics[width=\linewidth]{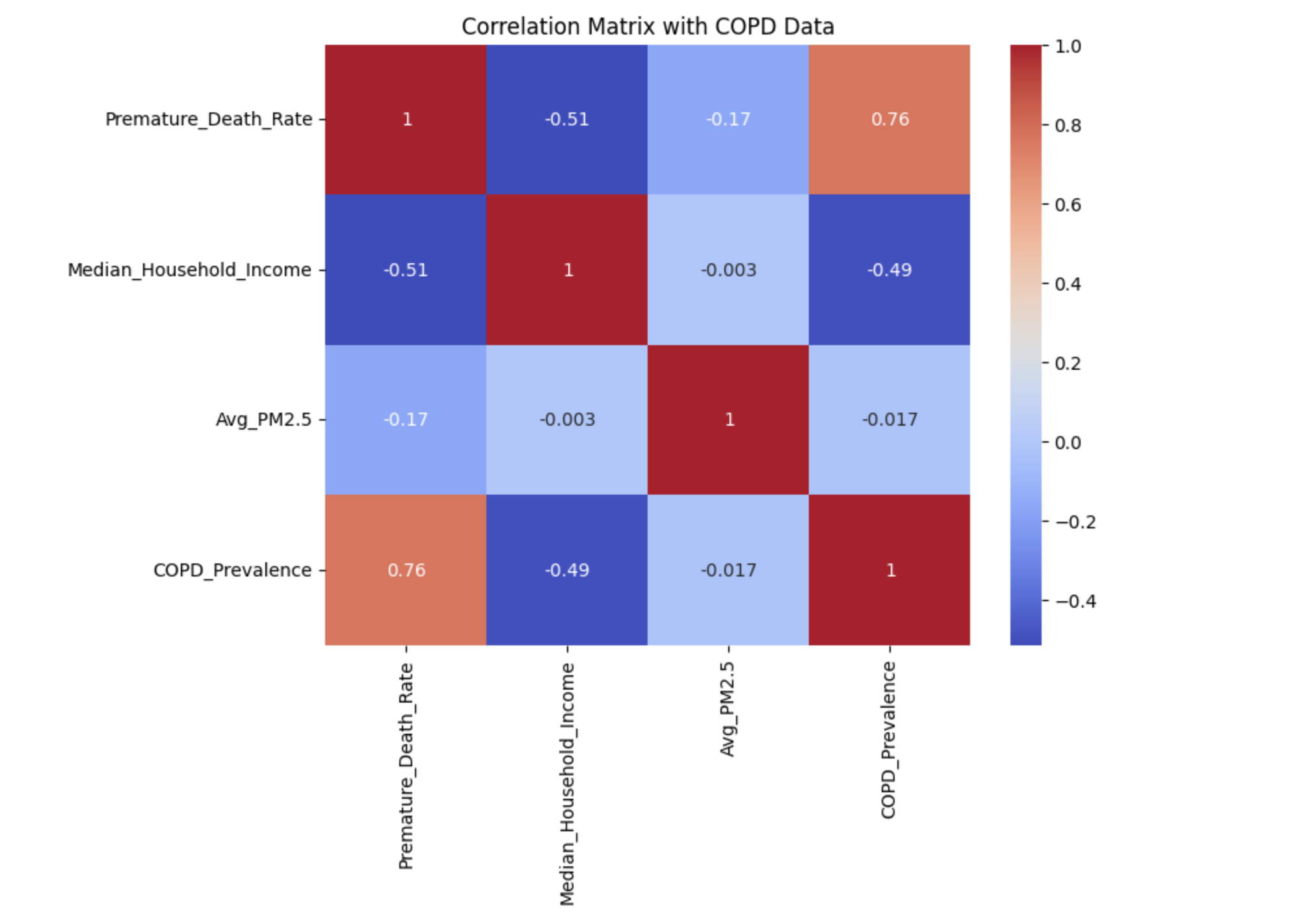}
\caption{Correlation Matrix of Key Variables.}
\label{fig:correlation_matrix}
\end{figure}

\subsection{Relationship Visualizations}

\begin{figure}[ht]
\centering
\includegraphics[width=\linewidth]{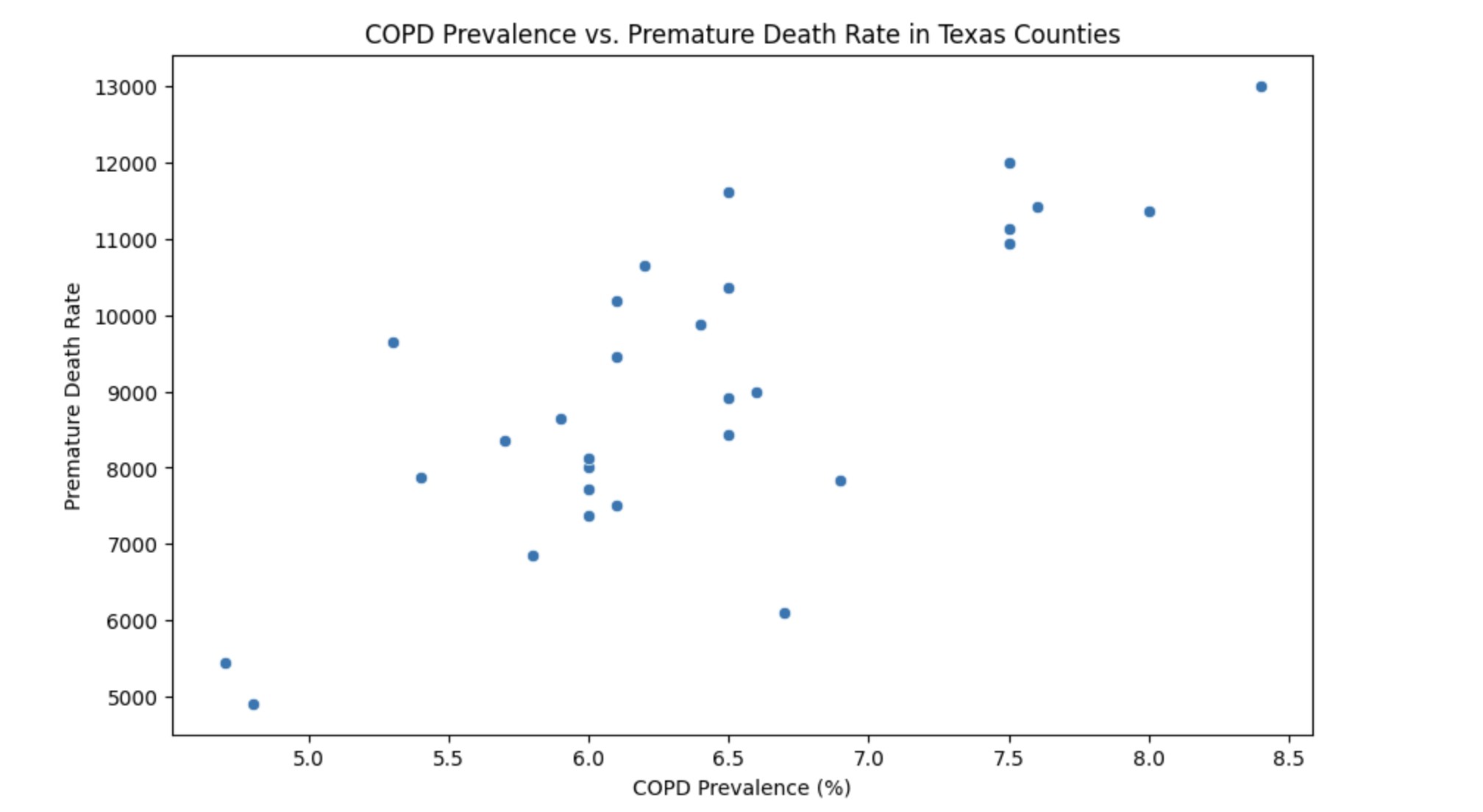}
\caption{COPD Prevalence vs. Premature Death Rate.}
\label{fig:copd_vs_premature_death}
\end{figure}

\begin{figure}[ht]
\centering
\includegraphics[width=\linewidth]{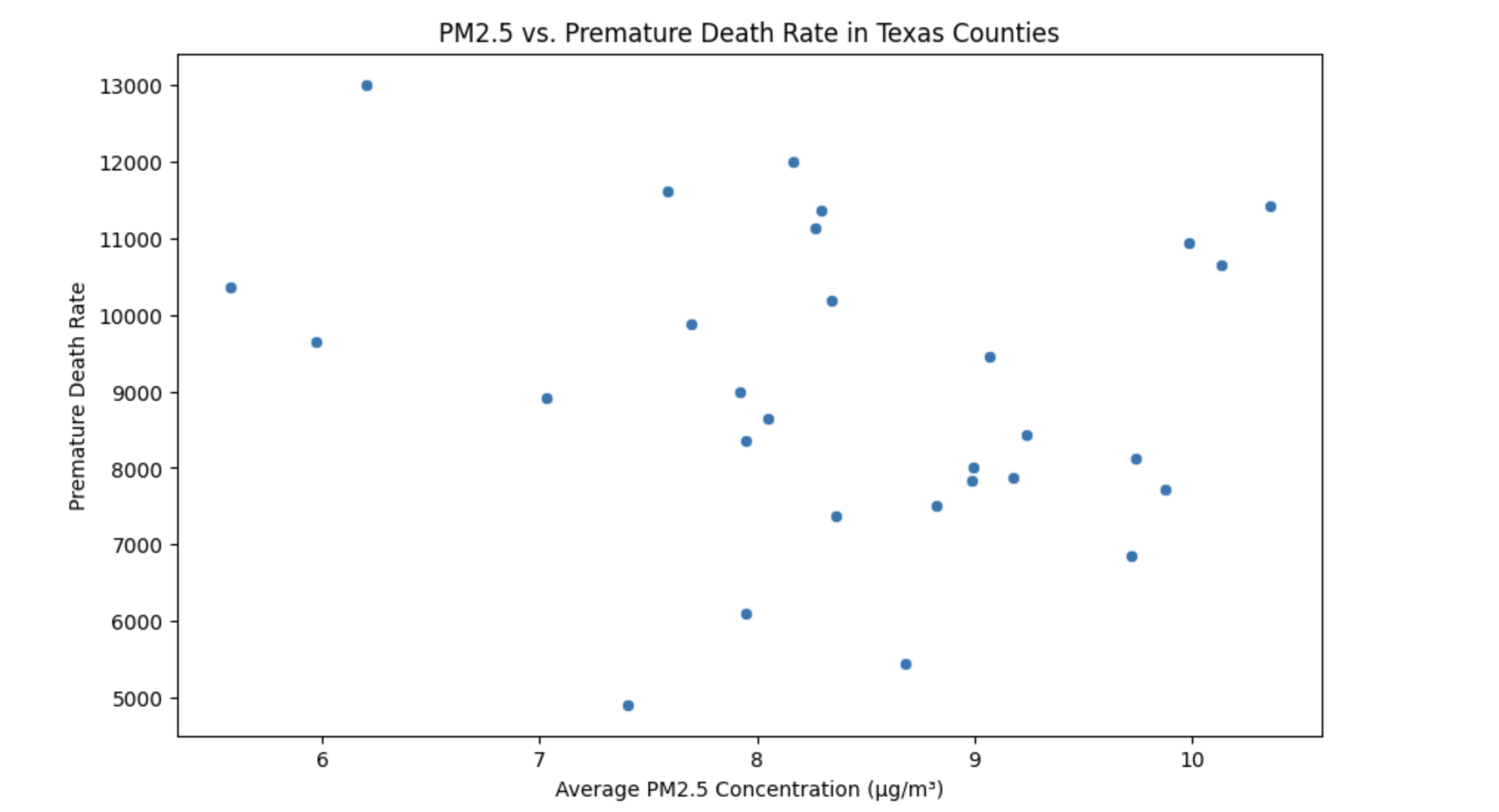}
\caption{PM2.5 vs. Premature Death Rate in Texas Counties.}
\label{fig:pm25_death}
\end{figure}

\begin{figure}[ht]
\centering
\includegraphics[width=\linewidth]{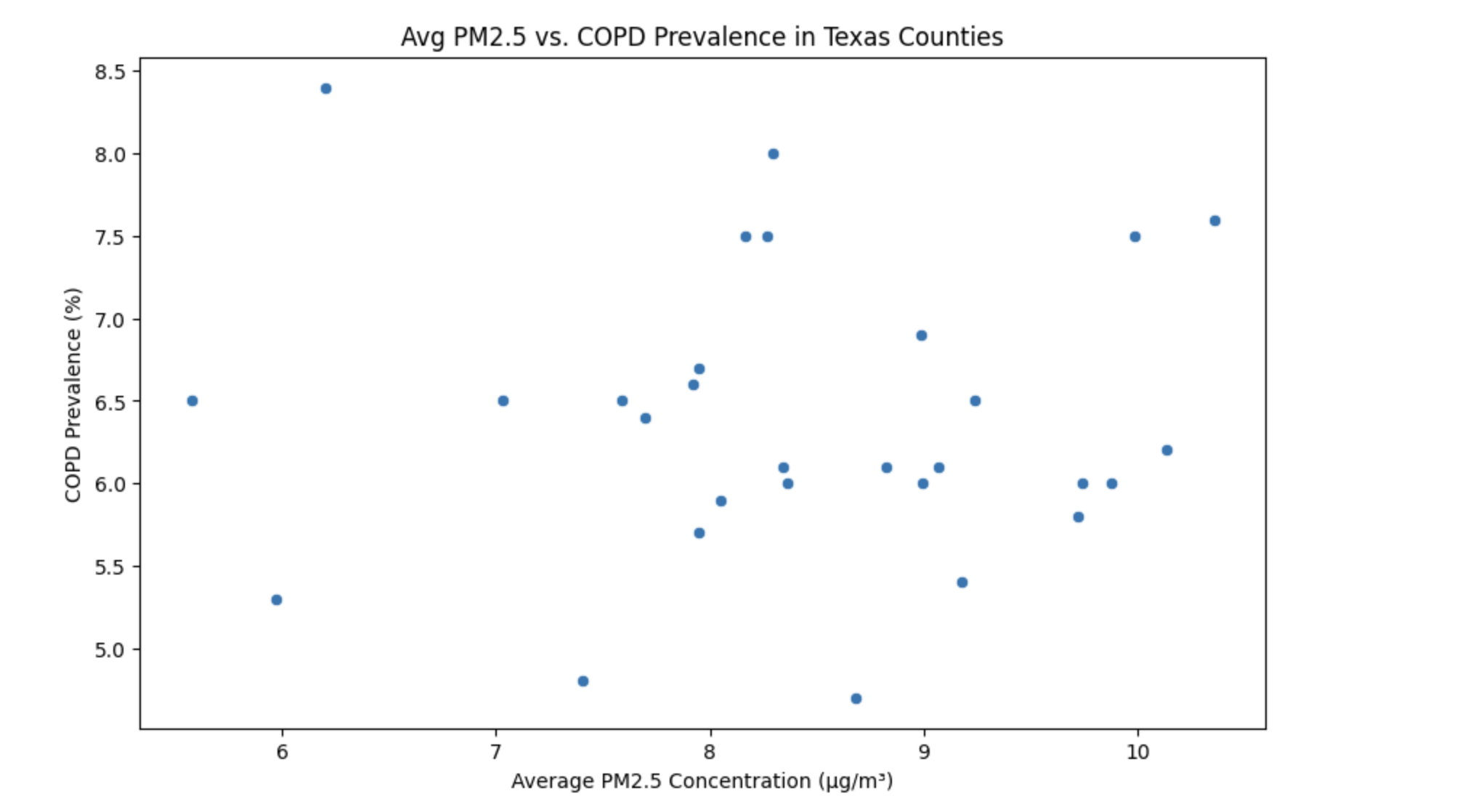}
\caption{Average PM2.5 vs. COPD Prevalence in Texas Counties.}
\label{fig:pm25_copd}
\end{figure}

The scatter plots illustrate the relationships between key variables. Figure \ref{fig:copd_vs_premature_death} demonstrates the strong positive correlation between COPD prevalence and premature death rates. Figure \ref{fig:pm25_death} shows a less clear relationship between PM2.5 levels and premature death rates, while Figure \ref{fig:pm25_copd} reveals minimal correlation between PM2.5 concentrations and COPD prevalence, suggesting complex underlying interactions between these health and environmental factors.

\section{Modeling Results}

\subsection{Linear Regression Model}
The model predicts premature death rates using the following equation:
\begin{equation}
\begin{split}
\text{YPLL} &= \beta_0 + \beta_1 \cdot \text{Income} + \beta_2 \cdot \text{PM}_{2.5} \\
           &\quad + \beta_3 \cdot \text{COPD Prevalence} + \epsilon
\end{split}
\end{equation}

\subsection{Model Performance}
\begin{itemize}
    \item Mean Absolute Error (MAE): 931.99
    \item Root Mean Squared Error (RMSE): 1,203.02
    \item \(R^2\): 0.81
\end{itemize}

\subsection{Feature Coefficients}
\begin{table}[ht]
\centering
\caption{Regression Coefficients}
\label{table:coefficients}
\begin{tabular}{lcc}
\toprule
\textbf{Feature} & \textbf{Coefficient} & \textbf{Effect} \\
\midrule
Median Household Income & -0.0277 & Negative \\
PM2.5 Levels & -234.22 & Negative \\
COPD Prevalence & 1,376.89 & Positive \\
\bottomrule
\end{tabular}
\end{table}

\subsection{Model Fit Visualization}
\begin{figure}[ht]
\centering
\includegraphics[width=\linewidth]{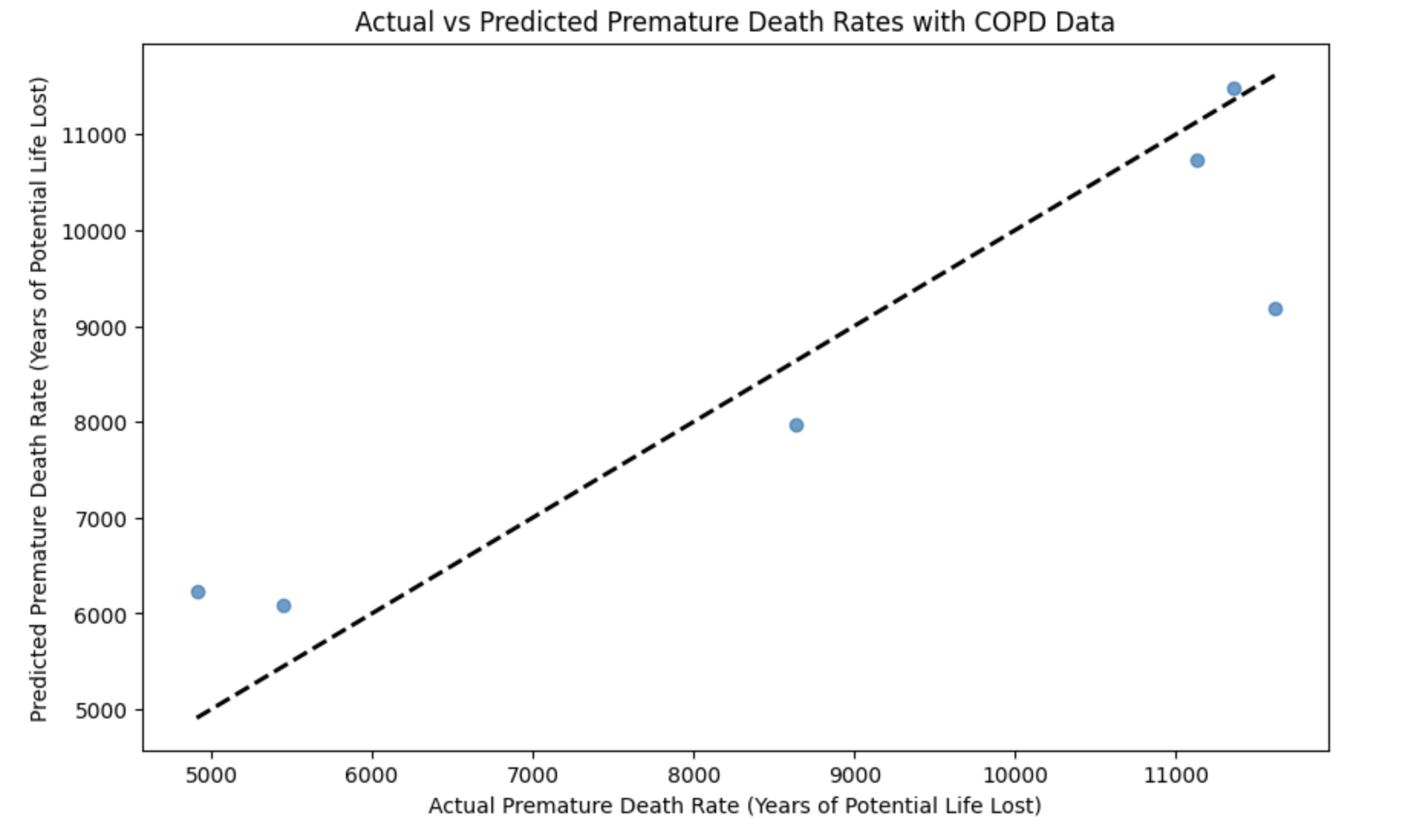}
\caption{Actual vs. Predicted Premature Death Rates.}
\label{fig:model_fit}
\end{figure}

\section{Discussion}
The analysis reveals several significant relationships between health outcomes and environmental/socioeconomic factors in Texas counties. COPD prevalence emerges as the strongest predictor of premature death rates (correlation: 0.76), with each percentage increase associated with approximately 1,377 additional years of potential life lost. This robust relationship underscores the critical importance of respiratory health management in public health strategies.

The substantial negative correlation (-0.51) between median household income and premature mortality rates highlights how socioeconomic factors influence health outcomes. This finding suggests that interventions addressing economic disparities could significantly impact public health, potentially through improved healthcare access and living conditions.

Recent large-scale studies have demonstrated the significant health impacts of PM2.5 exposure across the United States, with disproportionate effects on socioeconomically disadvantaged communities. A comprehensive study of over 4.5 million US veterans found that 99\% of PM2.5-associated deaths occurred at levels below current EPA standards, with higher burden among those in areas of high socioeconomic deprivation \cite{bowe2019burden}. Our analysis of Texas counties similarly reveals concerning relationships between environmental factors and mortality rates, particularly in relation to socioeconomic status.
While PM2.5 levels show a weaker direct relationship with mortality than expected, this finding warrants further investigation. The complex interactions between air quality, chronic health conditions, and socioeconomic factors suggest that PM2.5's influence may operate through indirect pathways. Additional research incorporating longitudinal data and more environmental variables could help clarify these relationships.

These findings collectively emphasize the need for a multi-faceted approach to reducing premature mortality, incorporating both direct health interventions and broader socioeconomic policies.

\section{Conclusion}
This study highlights the complex interplay between health conditions, socioeconomic status, and environmental factors in determining premature mortality rates across Texas counties. COPD prevalence emerged as a dominant predictor, while household income demonstrated significant protective effects, emphasizing the need for integrated public health strategies. Future research could expand this analysis through longitudinal studies, and additional health indicators to better inform targeted interventions and policy decisions.

\section*{Acknowledgments}
The authors thank the U.S. Census Bureau, EPA, and County Health Rankings for providing open-access data.

\bibliographystyle{ieeetr}

\end{document}